\documentclass[twoside,11pt]{article}

\usepackage{blindtext}

\usepackage{jmlr2e}

\usepackage{lastpage}
\ShortHeadings{SemanticStitcher}{SemanticStitcher}
\firstpageno{1}

\begin{document}

\title{Semantic Mosaicing of Histo-Pathology Image Fragments using Visual Foundation Models}

\author{\name Stefan Brandstätter\textsuperscript{1,3,4} \email stefan.brandstaetter@meduniwien.ac.at \\
        \name Maximilan Köller\textsuperscript{5} \email maximilian.koeller@meduniwien.ac.at \\
        \name Philipp Seeböck\textsuperscript{1,3,4}  \email philipp.seeboeck@meduniwien.ac.at \\
        \name Alissa Blessing\textsuperscript{5} \email alissa.blessing@meduniwien.ac.at \\
        \name Felicitas Oberndorfer\textsuperscript{3,5} \email felicitas.oberndorfer@meduniwien.ac.at \\
        \name Svitlana Pochepnia\textsuperscript{2,3}   \email svitlana.pochepnia@meduniwien.ac.at \\
        \name Helmut Prosch\textsuperscript{2,3}  \email helmut.prosch@meduniwien.ac.at \\
        \name Georg Langs\textsuperscript{1,3,4}  \email georg.langs@meduniwien.ac.at \\
        \addr \textsuperscript{1}Computational Imaging Research Lab \\
        %Department of Biomedical Imaging and Image-guided Therapy \\
        %Medical University of Vienna\\
        %Vienna, Austria \\
        \addr \textsuperscript{2}Division of General and Pediatric Radiology \\
        %Department of Biomedical Imaging and Image-guided therapy\\
        %Medical University Vienna\\
        %Vienna, Austria \\
        \addr \textsuperscript{3}Christian Doppler Laboratory for Machine Learning Driven Precision Imaging \\
        %Department of Biomedical Imaging and Image-guided therapy\\
        %Medical University Vienna\\
        %Vienna, Austria \\
        \addr \textsuperscript{4}Comprehensive Center for Artificial Intelligence in Medicine\\
        %Department of Biomedical Imaging and Image-guided therapy\\
        %Medical University Vienna\\
        %Vienna, Austria \\
        \addr \textsuperscript{5}Department of Pathology\\
        %Department of Biomedical Imaging and Image-guided therapy\\
        Medical University Vienna, Vienna, Austria \\ 
        }
\editor{}
\maketitle
\begin{abstract}%   <- trailing '%' for backward compatibility of .sty file
In histopathology, tissue samples are often larger than a standard microscope slide, making stitching of multiple fragments necessary to process entire structures such as tumors. Automated stitching is a prerequisite for scaling analysis, but is challenging due to possible tissue loss during preparation, inhomogeneous morphological distortion, staining inconsistencies, missing regions due to misalignment on the slide, or frayed tissue edges. This limits state-of-the-art stitching methods using boundary shape matching algorithms to reconstruct artificial whole mount slides (WMS). Here, we introduce SemanticStitcher using latent feature representations derived from a visual histopathology foundation model to identify neighboring areas in different fragments. Robust pose estimation based on a large number of semantic matching candidates derives a mosaic of multiple fragments to form the WMS. Experiments on three different histopathology datasets demonstrate that SemanticStitcher yields robust WMS mosaicing and consistently outperforms the state of the art in correct boundary matches.
\end{abstract}


%%%%%%%%%%%%%%%%%%%%%%%%%%%%%%%%%%%%%
% INTRODUCTION SECTION
%%%%%%%%%%%%%%%%%%%%%%%%%%%%%%%%%%%%%
\section{Introduction}
While microscope slides are essential for pathology, their size limits full specimen analysis.~\citep{Duan2024-rw} Whole-Mount histopathology addresses this but introduces new scanning challenges. Artificial WMS mosaicing offers a solution through fragment alignment.
Whole-Mount histopathology (WMH) is a comprehensive technique examining the entire cross-section of a specimen resulting in a Whole-Mount Sectioning (WMS) large-format slide.~\citep{Cimadamore2020-ug} It captures the full spatial distribution and morphological features of tissue relevant for diagnosis and research. WMS enhances histopathology-imaging correlation, reduces cutting artifacts, and preserves tissue context.~\citep{Schouten2024-mv} However, it also introduces challenges, including the need for larger, more costly scanners, and technical limitations in capturing oversized slides.~\citep{Duan2024-rw}
To address these limitations, ongoing research focuses on creating artificial WMS images by aligning tissue fragments, to obtain advantages of WMS while maintaining standardized image acquisition protocols.~\citep{CHAPPELOW2011557,autostitcher_cite,Schouten2024-mv}

\paragraph{Related work}
%HistoStitcher
Several stitching methods have been proposed. HistoStitcher~\citep{CHAPPELOW2011557} requires manual landmark selection and transformation tuning (e.g., reflection, scaling), making it too labor-intensive for clinical use
%Several approaches for stitching have been introduced. HistoStitcher~\citep{CHAPPELOW2011557} relies on manually selecting landmarks and specifying transformations (e.g., reflection, scaling) to compute an optimal transformation minimizing keypoint errors. Its dependency on expert input and time-intensive process renders it unsuitable for routine clinical use.
%AutoStitcher
AutoStitcher~\citep{autostitcher_cite} was the first fully automated method, using a domain-specific cost function based on L2-norm histogram differences and a misalignment term, but still depends on manual fragment labels. 
%AutoStitcher~\citep{autostitcher_cite} was the first fully automated approach. It incorporates a domain-specific cost function based on the L2-norm differences of histogram vectors and a misalignment term. However, it reliance on manual fragment labels. 
%PythoStitcher
PythoStitcher~\citep{Schouten2024-mv} is the current state-of-the-art, applying a boundary-based, multi-resolution strategy for high-resolution mosaicing without manual input or extra cost functions. However, it performs poorly on irregularly shaped or equally sized boundary fragments. To address these limitations while retaining WMS benefits, we propose a novel automatic mosaicing method
%PythoStitcher~\citep{Schouten2024-mv} represents the state-of-the-art, using a boundary-based, multi-resolution alignment strategy to achieve high-resolution mosaicing without manual input or additional cost functions. Although effective for prostate histological slides fragmented into four pieces, it struggles with arbitrary-shaped tissues divided more equal-length boundaries fragments.
To take advantage of WMS while addressing associated challenges, we propose a novel method for automatic mosaicing of artificial WMS from given tissue fragments. %This approach utilizes semantic-level stitching powered by a foundation model. 

 %2 saetze zur novelty der methodik
 % 2 saetze zu den beneficial konsequenzen verglichen mit dem state of the art
 %evtl. ein kommentar, dass das zeigt, das foundation models fuer semantic features gut sind, was speziell wichtig ist bei histo daten, weil die original bilddaten so hochaufgeloest sind
\paragraph{Contribution} We introduce SemanticStitcher, an automated WMS mosaicing method that aligns arbitrary boundary fragments using direct image content stitching. Unlike state-of-the-art boundary-based algorithms, it requires no prior knowledge of tissue shape, arrangement, or fragment count. Instead, it leverages semantic features extracted with the help of a foundation model to effectively compare the content of image patches at high resolution, enabling robust and accurate alignment of fragment boundaries.

%%%%%%%%%%%%%%%%%%%%%%%%%%%%%%%%%%%%%
% METHOD SECTION
%%%%%%%%%%%%%%%%%%%%%%%%%%%%%%%%%%%%%
\section{Method}
\begin{figure}[!t]
\includegraphics[width=\linewidth]{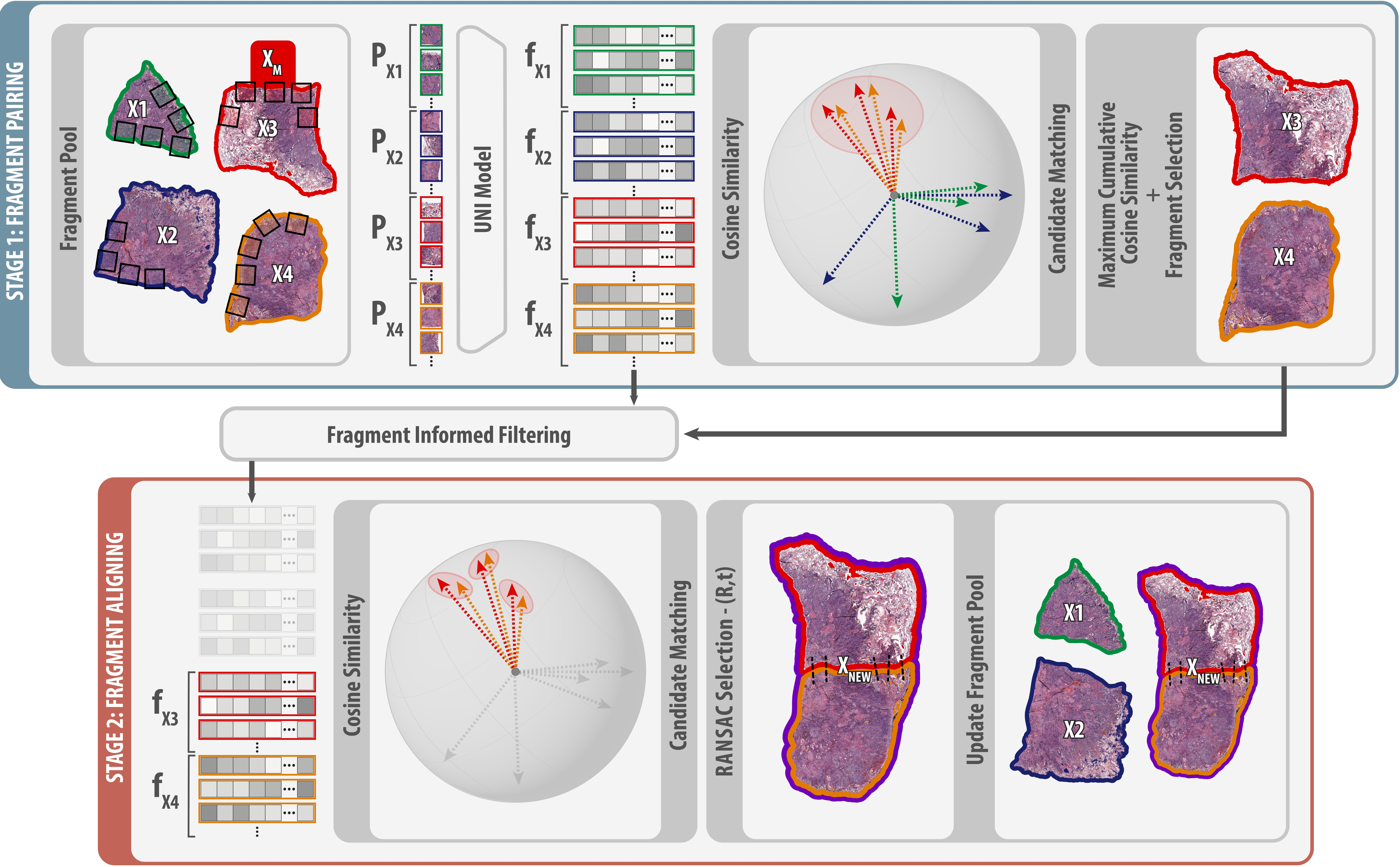}
\caption[Grafical abstract]{\textbf{SemanticStitcher:} In Stage 1, a fragment is randomly selected from the fragment pool and paired with its most compatible counterpart.
In Stage 2, the optimal transformation aligning the paired fragments is computed.}
\label{fig:grafical_abstract}
\end{figure}

We propose \textit{SemanticStitcher}, a method for accurately aligning digitized tissue fragments \(\mathbf{X}\in\mathbb{R}^{h\times w}\) to reconstruct an artificial WMS (Fig.~\ref{fig:grafical_abstract}).
%\(\mathbf{X}_M\)\(\mathbf{X}_F\)
The algorithm selects consecutive fragments from a fragment pool, pairs it with its best-matching counterpart, and computes the rotation matrix $\mathbf{R}$ and the translation vector $\mathbf{t}$ required for precise alignment of the fragment.
The approach consists of two stages:  (1) Identifying neighboring fragment pairs and corresponding semantic match candidates and (2) robustly estimating their spatial alignment for mosaicing.

\paragraph{\textbf{Stage 1: Fragment pairing}}
As a simple preprocessing step, for each fragment $\mathbf{X}$ in the fragment pool we remove the background (OTSU~\citep{4310076}) and detect the boundary  \(\mathcal{B}_X\)~\citep{SUZUKI198532}. Along the fragment boundaries, patches \(\smash{\mathbf{P}_{X}^{(k)} \in \mathbb{R}^{ph\times pw}} \) are sampled at fixed intervals. Each patch is encoded into a feature vector \(\smash{\mathbf{S}:\mathbf{P}_{X}^{(k)}\mapsto\mathbf{f}_{X}^{(k)} \in \mathbb{R}^{K}}\) using a pre-trained semantic encoding model. In our experiments we evaluate the UNI and CONCH foundation model~\citep{chen2024uni, Lu_2023_CVPR}. 
A random fragment \(\mathbf{X}_M\) out of the pool is selected as the moving fragment. All other fragments are treated as fixed fragments $\mathbf{X}_{F}$. To determine the best fragment match, we compute the cosine similarity between all feature vectors of the moving fragment \(\mathbf{X}_M\) and all boundary patch feature vectors of all fixed fragments $\mathbf{X}_{F}$. For each feature vector in the moving fragment $\smash{\mathbf{f}_{X_M}^{(k)}}$, we identify the most similar feature vector on the fixed fragments $\smash{\mathbf{f}_{X_F}^{(j)}}$, forming candidate matches.
To select a matching fragment, we sum up the cosine similarity for each \(\mathbf{X}_F\) and select the fixed fragment with the highest score as the optimal match for \(\mathbf{X}_M\).

\paragraph{\textbf{Stage 2: Fragment alignment }} To align the fragments, we first apply fragment informed filtering, where only the feature vectors of the matched fragment pair \((\mathbf{X}_M, \mathbf{X}_F)\) from Stage 1 are retained for alignment. We then identify candidate matches between \(\mathbf{X}_M\) and \(\mathbf{X}_F\) by finding feature pairs with the highest similarity. The candidate matches form the basis for a robust estimation of alignment parameters between the fragments using RANSAC~\citep{10.1145/358669.358692}. It mitigates the impact of outliers and incorrectly matched pairs, and identifies the pose parameters consistent with a majority of matches. Once \(\mathbf{X}_M\) is aligned with \(\mathbf{X}_F\), the resulting composite image replaces the two initial fragments in the fragment pool, and the fragment selection and alignment is repeated. This iterative process continues until a single fragment remains, forming the complete WMS. In the following we provide details of the individual steps of the approach. 

%\paragraph{\textbf{Boundary extraxtion}} 
%To extract the boundary  $\mathbf{B}_X$ of a fragment, we first apply OTSU~\cite{4310076} thresholding to distinguish the foreground from the background. The contour of the fragment is then detected using the findContours function of OpenCV~\cite{opencv_library}.

%\subsection{Method Details}

\paragraph{\textbf{Patch extraction}} 
For patch extraction, in our experiments, we begin with a boundary point and identify the next boundary point at a distance of 224 pixels, corresponding to the input size of the UNI encoder. The patch \(\smash{\mathbf{P}_{X}^{(k)}}\) is then extracted perpendicular to the line connecting these two points. To ensure comprehensive coverage of the boundary $\mathcal{B}_X$, patches are extracted with an overlap of half the patch size. Patch extraction is shifted 10 pixels inward to avoid black pixels from frayed edges. This process is performed along the entire extracted boundary.

\paragraph{\textbf{Patch encodings}} 
To encode the patches \(\smash{\mathbf{P}_{X}^{(k)}}\) into meaningful feature vector representations \(\smash{\mathbf{f}_{X}^{(k)}}\), we utilize the state-of-the-art foundation model UNI, a general-purpose, self-supervised model for pathology.~\citep{chen2024uni} We follow the preprocessing described in the UNI paper.

%For each feature $\mathbf{f}_{I_M}$, we consider a neighborhood of three adjacent features and calculate the cosine similarity to all other features in the fixed images as a sliding window approach. 
\paragraph{\textbf{Feature vector matching}} 
To establish correspondences between the moving fragment \(\mathbf{X}_M\) and fixed fragment \(\mathbf{X}_F\), we consider both local patch correspondences and broader spatial relationships. For each patch $\smash{\mathbf{P}_{X}^{(k)}}$, we extract its associated feature vector \(\mathbf{f}_{X}^{(k)}\) along with three preceding and three following feature vectors, forming a context-aware feature stack \(\smash{\mathbf{F}_{X}^{(k)} = [\mathbf{f}_{X}^{(k-3)}, \dots, \mathbf{f}_{X}^{(k)}, \dots, \mathbf{f}_{X}^{(k+3)}]}\). 
We then compute the cosine similarity between each stacked feature vector \(\smash{\mathbf{F}_{X_M}^{(k)}}\) from \(\mathbf{X}_M\) against candidate stacks \(\smash{\mathbf{F}_{X_F}^{(j)}}\) from \(\mathbf{X}_F\) using a sliding window approach. The highest-scoring pair for each \(\smash{\mathbf{F}_{X_M}^{(k)}}\) is selected, generating candidate matches.

\paragraph{\textbf{Robust Transformation Estimation}}
We use the candidate matches to calculate the transformation between the moving and fixed image using the RANSAC algorithm. RANSAC provides the most consistent rotation matrix ($\mathbf{R}$) and translation vector ($\mathbf{t}$), both of which are utilized for the final alignment of the moving and fixed image. 

%%%%%%%%%%%%%%%%%%%%%%%%%%%%%%%%%%%%%
% EXPERIMENTAL-SETUP SECTION
%%%%%%%%%%%%%%%%%%%%%%%%%%%%%%%%%%%%%
\section{Experimental Setup}
\textbf{Data}
We utilized three medical imaging datasets from two different organs. The first dataset, TCGA-LUAD~\citep{https://doi.org/10.7937/k9/tcia.2016.jgnihep5}, comprises 514 tissue slides of lung adenocarcinoma. The second dataset, TCGA-PRAD~\citep{https://doi.org/10.7937/k9/tcia.2016.yxoglm4y}, consists of 490 tissue slides of prostate adenocarcinoma. Both datasets were reduced to 310 and 254 samples, respectively, due to the presence of painted slides, insufficient resolution, and excessively frayed or torn tissue samples, rendering them unsuitable for analysis. We use them to artificially simulate different fragment arrangements, and evaluate the corresponding performance of the stitching algorithm. The third dataset is an in-house collection comprising 8 hematoxylin and eosin (HE) stained lung cancer specimens, scanned using an Olympus VS200 slide scanner, at resolution 0,274 \textmu m/px. This dataset comprises real fragments with corresponding irregularities representative of clinical procedures.
Experiment A was performed on the in-house dataset, Experiment B on all three datasets and Experiment C, D and E using the TCGA-LUAD and TCGA-PRAD datasets. 
\\
\textbf{Implementation details}
All slides were processed at a resolution of 1 \textmu m and reconstructed at 0.25 \textmu m.
For all analyses, we used the pretrained UNI model. Additionally, for experiment B, we incorporated the pretrained CONCH model~\citep{Lu_2023_CVPR}. For the UNI/CONCH model the extracted patches $\smash{\mathbf{P}_{X}^{(k)}}$ had a size of 224$\times$224/448$\times$448 and the feature vectors $\smash{\mathbf{f}_{X}^{(k)}\in\mathbb{R}^{K}}$ had a size of $K=1024$/$K=768$.
The parameters of the RANSAC~\citep{10.1145/358669.358692} algorithm were set as follows: an inlier threshold of 500 pixel, a maximum of 1000 iterations, and a minimum of 6 points required for model estimation.
%All experiments were performed on a system equipped with an AMD Ryzen 5 3600 processor and an NVIDIA GeForce RTX 2060 Super.

\noindent \textbf{Experiment A: WMS reconstruction in clinical practice}
We assess the applicability of our method in clinical practice on our in-house dataset of histological fragments generated during routine diagnostics. These fragments were reconstructed into an artificial WMS, and the arrangement was compared to the expert-provided ground truth arrangement. We compared SemanticStitcher to the state of the art PythoStitcher. 

\noindent \textbf{Experiment B: Quantitative evaluation of tissue alignment} 
We quantitatively evaluate the tissue alignment accuracy on all three datasets by (1) artificially splitting a tissue slide into fragments, (2) mosaicing the artificial fragments to a WMS, and (3) counting the number of correctly vs. incorrectly matched boundaries.

\noindent \textbf{Experiment C: Spatial awareness and impact of RANSAC}
We visualize the connections between patches in the moving fragment and their corresponding matches in the fixed fragment before and after RANSAC. Additionally, we qualitatively assess spatial and semantic awareness - or capture range - of the feature space by visualizing the cosine similarity between an encoded patch from the image center and all other tissue patches.

\noindent \textbf{Experiment D: Accuracy of correct match prediction with increased gap size}
We evaluate patch matching accuracy by artificially splitting a tissue slide into two fragments before and after RANSAC, and progressively increasing the gap between them. This simulates real-world conditions, where tissue gaps may be larger due to morphological distortions, misalignments on the slide, or inaccurate cuts during routine processing. We assess how patch distance affects feature embeddings via cosine similarity between patches with an offset varying between 0 and 900 $\mu m$.

\noindent \textbf{Experiment E: Rotation invariance, neighborhood analysis and resolution invariance}
We assessed the practicality of our method by conducting in depth analyses of (1) rotation invariance, to ensure that a rotated patch remains most similar to itself, (2) the effect of neighborhood size on feature matching, and (3) the trade-off between resolution and computational speed.

%%%%%%%%%%%%%%%%%%%%%%%%%%%%%%%%%%%%%
% RESULTS SECTION
%%%%%%%%%%%%%%%%%%%%%%%%%%%%%%%%%%%%%
\section{Results}
\begin{figure}[!t]
\includegraphics[width=\linewidth]{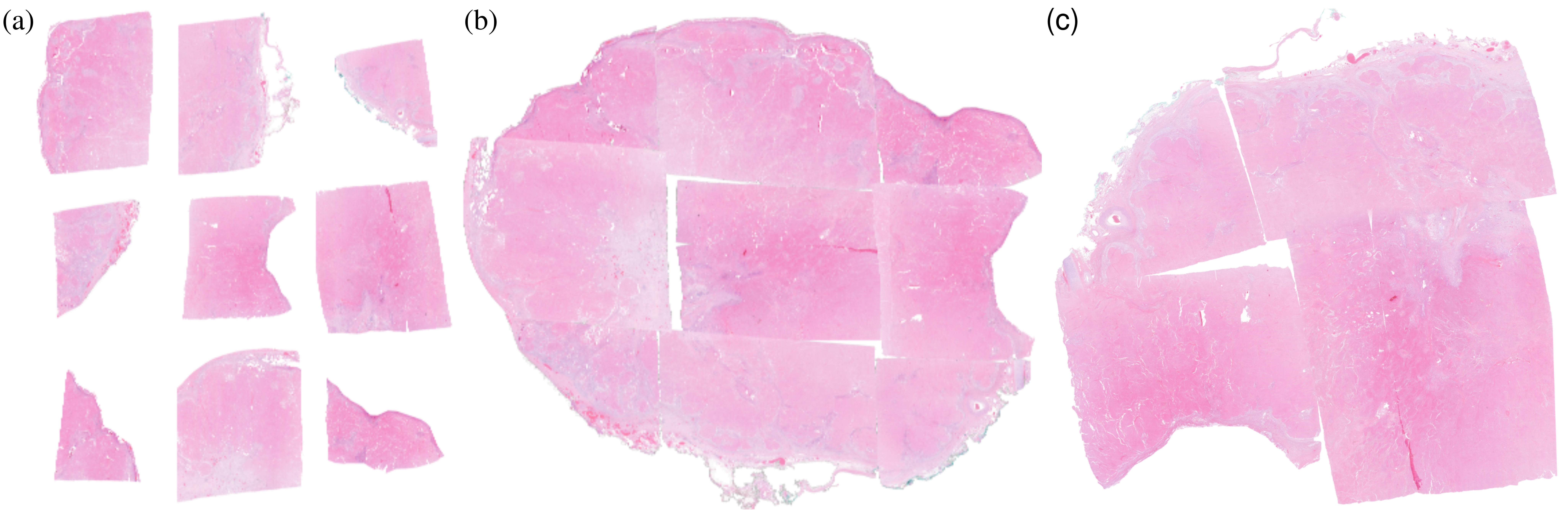}
\caption[qualitative results]{Experiment A: (a) Visualization of raw fragments without preprocessing, (b) WMS reconstruction using SemanticStitcher, (c) WMS reconstruction using the state‐of‐the‐art PythoStitcher algorithm}
\label{fig:qualitative_results}
\end{figure}

%%%%%%%%%%%%%%%%%%%%%%%%%%%%%%%%%%%%%
% EXPERIMENT A results
%%%%%%%%%%%%%%%%%%%%%%%%%%%%%%%%%%%%%
\noindent \textbf{A.} Fig.~\ref{fig:qualitative_results} shows qualitative results of real-world fragment mosaicing without any predefined orientation or arrangement. SemanticStitcher yields robust mosaicing results with all boundaries correctly matched, and only minor alignment errors observed in fragments with frayed boundaries (b top right segment) or where portions of the tissue were missing from the slide during imaging (b center segment). In contrast, we observed significant inaccuracies in fragment positioning and stitching edge alignment for the state-of-the-art boundary-based approach. It has difficulty processing fragments with similar-length boundaries (squares), as observed in (c) at the bottom left and right.

%To address overlapping tissues, we extend the alignment of the moving image to the fixed image, ensuring no overlapping tissue. Minor alignment errors were observed in fragments with frayed boundaries (b top right segment) or where portions of the tissue were missing from the slide during imaging (b center segment). Fig.~\ref{fig:qualitative_results} (c) presents the results of the state-of-the-art boundary-based approach, highlighting inaccuracies in fragment positioning and stitching edge alignment.

%%%%%%%%%%%%%%%%%%%%%%%%%%%%%%%%%%%%%
% EXPERIMENT B results
%%%%%%%%%%%%%%%%%%%%%%%%%%%%%%%%%%%%%
\noindent \textbf{B.} Table~\ref{sample-table} reports boundary matches in \%. Results demonstrate that SemanticStitcher consistently outperforms the state‐of‐the‐art algorithm across all three datasets by a large margin. For the in-house dataset, no preprocessing was applied, and both algorithms received raw fragments. For the TCGA-LUAD and TCGA-PRAD datasets, tissue slides were split into four same size segments to facilitate boundary matching. To simulate routinely scanned slides, we increased the gap between fragments to the size of one patch and randomly reduced the stitching edges by 0–20\% to mimic variable boundary lengths.

\begin{table}[ht]
\centering
\caption{Quantitative analysis of boundary matches (in \%).}
\label{sample-table}
\begin{tabular}{lccc}
\hline
\textbf{Method} & \textbf{TCGA-LUAD} & \textbf{TCGA-PRAD} & \textbf{IN-HOUSE} \\ 
&&Matches in \% $\uparrow$& \\
\hline
PythoStitcher~\citep{Schouten2024-mv}  & 42.21 & 46.12 & 38.88  \\
\textbf{SemanticStitcher (ours)}       & \textbf{81.33} & \textbf{76.05} & \textbf{86.11} \\
\hline
\end{tabular}
\end{table}

\begin{figure}[t!]
\includegraphics[width=\linewidth]{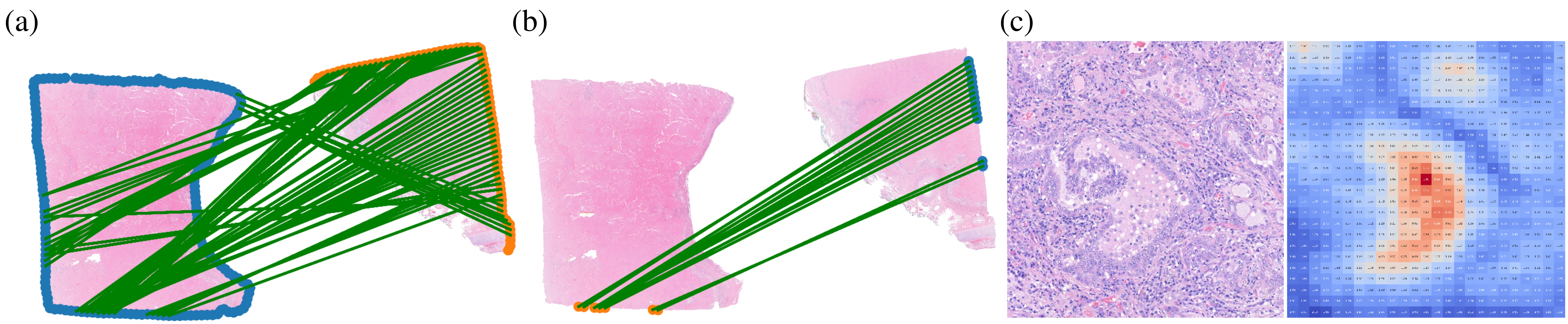}
\caption[Before after Ransac heatmap]{Experiment C: (a) Visualization of all semantic match candidates between two matched fragments; (b) valid connections retained after RANSAC filtering; (c) a tissue slide (left) and  the feature space cosine similarity between the central patch and the rest of the image (right).}
\label{fig:before_after_ransac_heatmap}
\end{figure}

%%%%%%%%%%%%%%%%%%%%%%%%%%%%%%%%%%%%%
% EXPERIMENT C results
%%%%%%%%%%%%%%%%%%%%%%%%%%%%%%%%%%%%%
\noindent \textbf{C.} Fig.~\ref{fig:before_after_ransac_heatmap}(a) displays a representative case for fragment matching (Stage 2), with all matched patches connected by lines. Several incorrect matches are observed, primarily due to factors such as staining variations, morphological distortions (e.g., tissue shrinkage), frayed tissue edges introducing black pixels during encoding, and slide misalignments. Fig.~\ref{fig:before_after_ransac_heatmap}(b) illustrates the correction of mismatches by applying the RANSAC algorithm. This is in line with quantitative results, demonstrating that erroneous connections can be effectively discarded by a consensus transformation across patches, omitting the need for perfect matches for every patch. Fig.~\ref{fig:before_after_ransac_heatmap}(c) illustrates the cosine similarity within the feature space by comparing the central patch's feature vector to those of all other patches in the image. It can be observed that patches closer to the center exhibit higher similarity, demonstrating that visually similar tissue regions also display similarity in the feature space.

\begin{figure}[th]

\includegraphics[width=\linewidth]{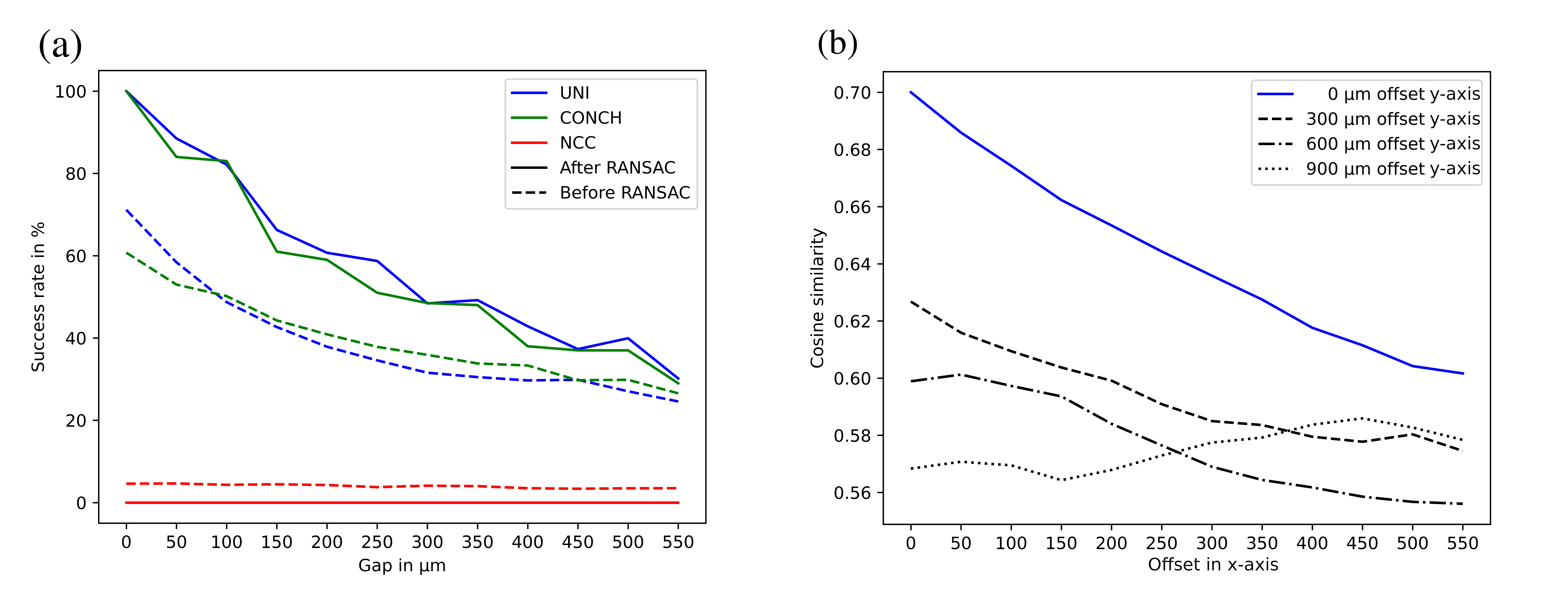}
\caption[Correct matches / similarity ]{Experiment D: (a) The fraction of correct matches of patches among all match candidates before and after RANSAC for different encoding models and a standard image-level similarity measure. (b) Cosine similarity of patch pair encodings with increasing tangential offset of pairs. (a) and (b) are evaluated as the gap between fragments is incrementally increased.}
\label{fig:correct_matches_similarity}
\end{figure}

%%%%%%%%%%%%%%%%%%%%%%%%%%%%%%%%%%%%%
% EXPERIMENT D results
%%%%%%%%%%%%%%%%%%%%%%%%%%%%%%%%%%%%%
\noindent \textbf{D.} Fig.~\ref{fig:correct_matches_similarity} (a) shows correct patch matches as percentage of all matches, before and after RANSAC, for the UNI model, its previously published counterpart CONCH, and a standard patch-level normalized-cross-correlation-based similarity (NCC) approach. Correct matches were determined using ground-truth fragments. Even at a 250 $\mu m$ gap—larger than a patch—SemanticStitcher matched over 40\% of cases without RANSAC.
%Correct matches are determined by comparison with known ground-truth matches in the artificially split fragments. RANSAC successfully removes incorrect matches, and as the gap increases, SemanticStitcher still correctly matches over 40\% of cases without using RANSAC, even when the gap reaches a relatively wide 250 $\mu m$, a distance greater than the size of one patch.
Fig.~\ref{fig:correct_matches_similarity} (b)  demonstrates that the cosine similarity of nearby patches is higher compared to patches further away. This indicates that using cosine similarity of feature embeddings both encodes spatial and semantic information. 
%Fig.~\ref{fig:correct_matches_similarity} (b) demonstrates the cosine similarity between patches with y-axis offsets of 0, 300, 600 and 900 $\mu m$. Additionally, the gap is incrementally increased. The results indicate that the cosine similarity between close patches consistently remains higher than that of the offset encoding pairs.

\begin{figure}[!t]
\includegraphics[width=\linewidth]{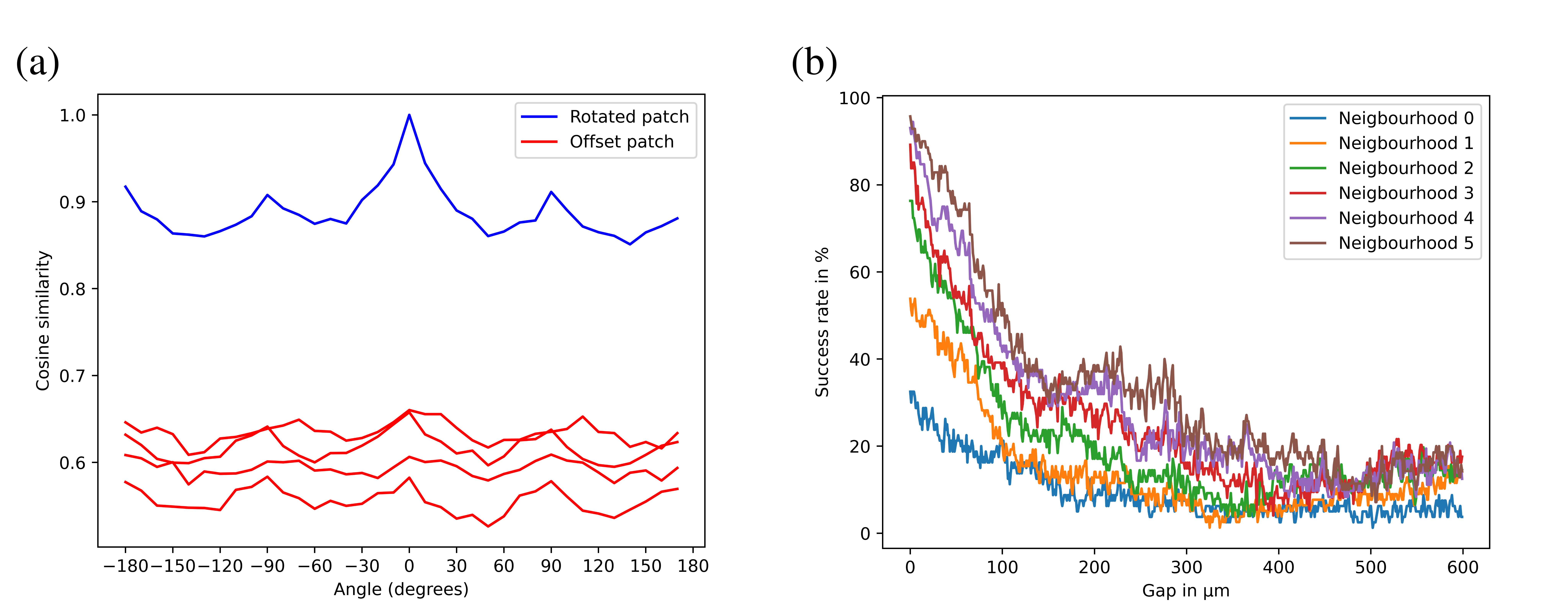}
\caption[Table keypoint error distance]{Experiment E: (a) Similarity between a rotated patch and its correct non-rotated counterpart (blue) compared to neighboring patches (red) illustrating the robustness of the comparison to rotations. (b) Success rate for matched patches for different neighborhood sizes as the gap size increases.}
\label{fig:deform_rotation_inv}
\end{figure}

%%%%%%%%%%%%%%%%%%%%%%%%%%%%%%%%%%%%%
% EXPERIMENT E results
%%%%%%%%%%%%%%%%%%%%%%%%%%%%%%%%%%%%%
\noindent \textbf{E.} Fig.~\ref{fig:deform_rotation_inv} (a) demonstrates rotation invariance of the embedding, with the rotated patch remaining more similar to its non-rotated counterpart than to four neighboring patches (top, right, left, and bottom). The observed peaks at 0°, 90°, and 180° are likely due to data augmentation used during the pretraining of UNI.
Fig.~\ref{fig:deform_rotation_inv} (b) highlights the importance of incorporating additional neighborhood information. The cosine similarity increases significantly from 30\% to 90\% when expanding from no neighborhood to a neighborhood size of 3, evaluated at a 0 $\mu m$ Fig.~\ref{fig:deform_rotation_inv} (b). We additionally evaluated SemanticStitcher across resolutions ranging from 0.25 to 4 $\mu m$ and identified 1 $\mu m$ as the optimal balance between efficiency and resolution.
%Fig.~\ref{fig:deform_rotation_inv} (c) demonstrates that SemanticStitcher performs well across various resolutions. However, an image resolution of 1 $\mu m$ achieved the highest success rate for gaps larger than 100 $\mu m$, which is particularly relevant in clinical routine scenarios where gaps are expected.

%%%%%%%%%%%%%%%%%%%%%%%%%%%%%%%%%%%%%
% DISCUSSION & CONCLUSION SECTION
%%%%%%%%%%%%%%%%%%%%%%%%%%%%%%%%%%%%%
\section{Discussion and Conclusion}
SemanticStitcher is a fully automated method for WMS mosaicing by aligning scanned fragments using semantic matching and robust pose estimation. Evaluation on three datasets shows its effectiveness on both synthetic and real clinical data. Unlike boundary-based methods, it leverages high-resolution semantic similarity, enabling more accurate and robust mosaicing. It captures neighborhood relationships despite boundary distortions from image acquisition. Compared to traditional metrics like cross-correlation, foundation models offer more stable representations for alignment. Though currently evaluated only on HE-stained slides, it is expected to generalize to other stainings with suitable embedding models. By handling boundary irregularities and diverse fragment layouts, SemanticStitcher can streamline workflows and enhance histopathology analysis.
\cite{xu2025topocellgengeneratinghistopathologycell}
%SemanticStitcher is a fully automated approach for reconstructing WMS by aligning multiple scanned fragments based on semantic matching and robust pose estimation. Evaluation on three datasets demonstrates its effectiveness on both synthetically fragmented and real-world clinical data. Unlike state-of-the-art methods that primarily rely on boundary matching, our approach relies on semantic similarity of high resolution imaging data, yielding more robust and accurate mosaicing results. It identifies neighborhood relationships despite boundary distortions that may occur during image acquisition. Compared to traditional similarity measures such as cross-correlation, visual foundation models provide robust representations for mosaicing histopathology images. It demonstrates that foundation models provide a sufficiently stable image characterization for correct alignment.  
%A limitation of our approach is that it was only evaluated on HE-stained slides. However, we expect our approach to perform well on other stainings given corresponding semantic embedding models. 
%By enabling more reliable WMS reconstruction regardless of boundary irregularities and varying fragment arrangements, the fully automatic SemanticStitcher can streamline pathology workflows and enhance histopathology analyses.
%. The results demonstrate the accurate reconstruction of artificial WMS using routinely acquired lung cancer tissue 

% Acknowledgements and Disclosure of Funding should go at the end, before appendices and references

\acks{The financial support by the Austrian Federal Ministry of Labour and Economy, the National Foundation for Research, Technology and Development,  the Christian Doppler Research Association, Siemens Healthineers, the Austrian Science Fund (FWF, P 35189-B - ONSET), the Vienna Science and Technology Fund (WWTF, PREDICTOME [10.47379/LS20065]), and the European Union’s Horizon Europe research and innovation programme under grant agreement No.101100633— EUCAIM are gratefully acknowledged.}
%\bibliography{sample}

\end{document}